\documentclass[letterpaper, 10 pt, conference]{ieeeconf}  

\IEEEoverridecommandlockouts                              
\usepackage{epsfig} 
\usepackage{amsmath} 

\title{\LARGE \bf
Sparser Sparse Roadmaps
}

\author{David Coleman and Nikolaus Correll$^{1}$
\thanks{DC was supported by NIST, SWRI.}
\thanks{$^{1}$David Coleman and Nikolaus Correll are with the Department of Computer Science,
        University of Colorado, Boulder, CO 80309, USA
        {\tt\small [david.t.coleman|nikolaus.correll]@colorado.edu}}%
}

\newtheorem{theorem}{Theorem}[section]
\newtheorem{lemma}[theorem]{Lemma}

\begin{document}

\maketitle
\thispagestyle{empty}
\pagestyle{empty}

\begin{abstract}

We present methods for offline generation of sparse roadmap spanners that result in graphs 79\% smaller than existing approaches while returning solutions of equivalent path quality. Our method uses a hybrid approach to sampling that combines traditional graph discretization with random sampling. We present techniques that optimize the graph for the $L_1$-norm metric function commonly used in joint-based robotic planning, purposefully choosing a $t$-stretch factor based on the geometry of the space, and removing redundant edges that do not contribute to the graph quality. A high-quality pre-processed sparse roadmap is then available for re-use across many different planning scenarios using standard repair and re-plan methods. Pre-computing the roadmap offline results in more deterministic solutions, reduces the memory requirements by affording complex rejection criteria, and increases the speed of planning in high-dimensional spaces allowing more complex problems to be solved such as multi-modal task planning. Our method is validated through simulated benchmarks against the ``SPARS2'' algorithm. The source code is freely available online \cite{bolt_github} as an open source extension to OMPL.

\end{abstract}

\section{Introduction}

Improving the query resolution time, path quality, and predictability of motion planning continues to be an important challenge today in resource-constrained applied robotics. We present the \textit{Bolt} algorithm that computes compact representations for shortest paths in continuous configuration spaces (c-spaces) in the form of roadmaps that are optimized for the $L_1$-norm metric space and that maintains asymptotically-near optimal theoretical guarantees on path quality. Bolt pre-processes the free c-space with invariant constraints such as self-collision checking, allowing faster online recall and repair of paths through changing collision environments. Such pre-processed roadmaps have applications for high dimensional dual arm robots, humanoids with balance constraints, and industrial applications where deterministic solutions are highly valued.

\begin{figure}[!t]
\centering
\includegraphics[width=3.4in]{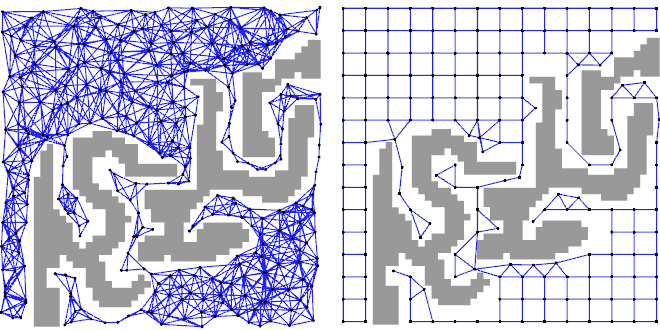}
\caption{Comparison of (left) SPARS2 roadmap: \textit{vertices: 334, edges: 1392} with (right) Bolt roadmap \textit{vertices: 173, edges: 262}. Both graphs return paths of the same quality in $L_1$ metric spaces.}
\label{fig:frontpage}
\end{figure}

Aiming at combining the best of graph search-based and sampling-based planning, Bolt uses a hybrid approach to generating a roadmap. We first apply a state space lattice pre-sampling step that inserts vertices into the graph at uniform increments of the c-space, followed by a random sampling step for filling in narrow passages. We discretize in joint-space rather than Cartesian work-space. The advantage of working in joint space is that it allows us to easily encode the redundancy in inverse kinematic solutions directly into the graph, with edges between vertices representing exact motions rather than under-defined end effector poses. It also allows us to avoid creating complex and specially-tuned heuristics typical of discrete graph-search planners such as \cite{aine2016multi}. The downside is the added dimensionality and space requirements of a larger c-space, the focus of this work. One advantage of our hybrid approach is that even coarse state space lattices can capture the necessary detail via the secondary random sampling step combined with asymptotically near-optimal quality path guarantees.

Choosing a proper distance function for joint-space (e.g. ${\rm I\!R}^6$) is non-intuitive due to the kinematic constraints of a robotic arm's geometry. Most literature has focused on SE3 spaces and variations of the $L_2$ Euclidean distance \cite{amato2000choosing}. Distance calculations are one of the most numerous operations in a PRM \cite{amato2000choosing} so for computational motivations the $L_1$-norm metric function (\textit{Manhattan distance}) is often used, such as in the MoveIt! Motion Planning Framework \cite{coleman2014reducing} that was used in this work. Using the $L_1$-norm will often return less smooth paths, but these can easily be smoothed in post-processing. Additionally, the $L_2$ distance between two points can be bounded by the $L_1$ distance. It is not clear that using $L_2$ or alternative distance functions \cite{amato2000choosing} are necessarily better than others, but the remainder of this work will assume $L_1$ is used for the computational and space advantages.

This work improves upon the SPARS \cite{spars_journal} graph spanners approach. Despite creating very sparse roadmaps, SPARS' original specification is still inefficient in its graph size for various reasons presented here. SPARS2 is ``slightly denser'' \cite{spars2} than the original SPARS1 algorithm, sacrificing graph density for lower initial graph construction memory requirements. Bolt is built upon SPARS2 \cite{spars2} and addresses those shortcomings, with the trade off of slower pre-processing times.  

Solution time in high dimension c-spaces such as dual arm robots is another goal of this work --- our planner searches the generated roadmap using the A* algorithm. A*'s time complexity is $O(|E|)=O(b^D)$, where $b$ is the \textit{branching factor} or average number of edges connected to each vertex, and $D$ is the depth of the search. Therefore reducing the number of edges in a graph will also reduce retrieval time for a path. Another often overlooked component for solving motion planning is the nearest neighbor search --- reducing the number of vertices results in important speedups.

Bolt was developed for use in experience-based planning --- the optimization of a planner for commonly used motion plans. As in its namesake ``Thunder'' \cite{coleman2015experience}, the goal is to remove the need for the planning from scratch component, instead relying fully on a pre-processed and highly efficient roadmap of the entire c-space. We reuse our roadmap multiple times not only for a given environment, but for all environments. Bolt can also be used for multi-modal task planning, allowing multi-goal motion planning problems to be quickly solved by cloning the entire joint-based Bolt roadmap for every discrete planning step.

\subsection{Related work}

One of the most popular methods for solving the robotic motion planning problem has been the sampling-based probabilistic roadmap (PRM) \cite{prm}. While in theory PRMs are pre-processed and reusable across all environments, in practice they are typically only re-usable for a given environment, though in some implementations more general re-use is actually achieved such as in Dynamic Roadmaps \cite{drm_kunz}. Even for versions of PRMs that can re-use their roadmap in changing environments, there are open questions on how to generate efficient and high quality graphs. Traditional PRMs are probabilistically complete but do not provide any guarantees on the quality of the path returned. PRM* \cite{prm_star} has been proven to be asymptotically optimal as samples are infinitely added to the roadmap. 

Sparse roadmap spanners, instead, have recently been proven to provide asymptotically near-optimal guarantees within a $t$-stretch factor. For example, if the $t$-stretch factor is 1.1, then the maximum length a path can be from its optimal solution is 10\%. The SPARS algorithm is powerful in that it is probabilistically complete, asymptotically near-optimal, and the probability of adding new vertices and edges to the roadmap converges to zero as new samples are added. This addresses the problem of unbounded growth suffered by PRM*, or the inability to only remove edges but not vertices \cite{marble2011asymptotically}. SPARS uses graph spanners to create subgraphs that approximate the roadmap PRM* would compute. This subgraph allows theoretical guarantees on path quality to be upheld while filtering out unnecessary vertices and edges from being added to the graph. 

In order to have the same asymptotic optimality guarantees as PRM* within a $t$-stretch factor, a number of checks are required to determine which potential vertices and edges should be saved to have coverage across a robot's free space. Only configurations that are useful for 1) coverage, 2) connectivity, or 3) improving the quality of paths on the sparse roadmap relative to the optimal paths in the c-space. Two parameters $t$ and the sparse delta factor $\Delta$ control the sparsity of the graph. For more background on these criteria the reader is encouraged to reference \cite{spars2}. 

An improved version, called SPARS2, relaxes the requirement for a dense graph to be maintained alongside the sparse graph. This greatly reduces the upfront memory requirements of the graph, making it practical for high DOF c-spaces to be maintained in memory. A small trade-off in graph density and new path insertion time is made for this relaxation, through a local sampling process and some bookkeeping information. Still, SPARS2 significantly reduces the size required for a roadmap with an asymptotically near-optimal guarantee.

Our hybrid approach of combining discretized lattices with random sampling is similar to the  extensive work on the subject in \cite{lavalle2004relationship}. They surprisingly found that deterministic sampling methods are superior the original PRM, noting that by definition a ``collection of pseudo-random samples should have too many points in some places, and not enough in others.'' Our approach is most similar to their proposed subsampled grid search (SGS), where discretized vertices along a grid are coarsely spaced and a local planner is used to collision check the edges between the grid. The unique aspect of our approach is that we size the grid optimally for the requirements of the spanning graph, and additionally perform random sampling as a second step. To the best of our knowledge this hybrid approach is unique in the literature.

\subsection{Contribution of this paper}

\textit{Bolt} is a new highly efficient sparse roadmap generator and re-use planner presented in this paper that has been utilized for motion planing in various c-spaces. Six modifications to the original method of SPARS are presented that result in an average roadmap size reduction of 79\% in 3D and an average planning time speedup of 35\%. These improvements include (and their corresponding edge reduction):

\begin{itemize}

\item A discretization pre-sampling step to efficiently cover large areas of free space in the c-space - 13\%.

\item An exact method for choosing the $t$-stretch factor optimized for the $L_1$ metric space - 15\%.

\item Methods to reduce outdated and redundant edges - 40\%.

\item An extra check for the connectivity criteria to determine if a vertex addition can be avoided - 36\%.

\item An additional smoothed quality path criteria - 12\%.

\item Modification of quality criteria for $L_1$ norm - 27\%.

\end{itemize}

We formally describe the proposed method and demonstrate their performance in a variety of 2D and 3D environments. 

\section{Methods For Creating Sparser Sparse Graphs}


We wish to create a sparse graph $G_S = (V, E)$ where $V$ is a set of points $q$ in the collision-free subset of the $d$-dimensional c-space $C_{free}$. Edges $E$ in $G_S$ have local paths $L(q_1,q_2)$ between two points in $V$. The local planner used in this work is the straight-line interpolation between to points.  The generated $G_S$ is then used to solve the \textit{path planning problem} for finding the shortest path between start and goal points $q_{start}, q_{goal} \in C_{free}$. This shortest path $\pi$ is a continuous path $\{q|q:[0,1] \rightarrow C_{free}\}$ where $\pi(0)=q_{start}$, $\pi(1)=q_{goal}$. The $L_1$-norm metric function $d(q_1,q_2)\rightarrow {\rm I\!R}$ is used in $C$ to find the distance between two configurations. In this work our cost function is always the shortest path, and our sampling is terminated after the number of failed samples is $M$.

The generated $G_S$ shall be \textit{asymptotically near-optimal with additive cost} if, for a cost function $c$ with an optimal path of finite cost $c^*$, the probability a path will be found with cost $c \leq t \cdot c^* + \varepsilon$ for a stretch factor $t \geq 1$ and additive error $\varepsilon \geq 0$, converges to 1 as number of samples approach infinity \cite{spars_journal}.

\subsection{Hybrid Discretization and Sampling}

Here we present the hybrid approach to generating a roadmap that combines graph search-based planning and sampling-based planning. Discrete graphs avoid redundancy by equally spacing vertices and edges through the c-space, but can miss narrow passages due to resolution coarseness or consume too much memory and search time. Whereas sampling-based planners address these issues, they are not efficient if the goal is to create a graph that provides the near-asymptotically optimal properties in the fewest vertices and edges possible. 

In Bolt, we initially cover the c-space with a discrete graph, providing efficient coverage of free space, and then sample in the more complex areas of space that interface with invalid regions. We must ensure that no vertex or edge added to the graph is in violation of the near-asymptotically optimal guarantees. In fact with the following discretization method, and in the absence of any obstacles or other constraints, the random sampler with SPARS criteria is unable to add any extra vertices beyond those added by the discretization step.

\begin{figure}[!t]
\centering
\includegraphics[width=3.4in]{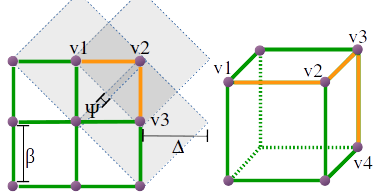}
\caption{Demonstration of how the optimal discretization factor $\beta$ is chosen for the $L_1$ norm in a 2D discretized grid (left) and 3D discretized grid (right). The gray transparent diamonds around the vertices $v1$, $v2$, $v3$ represent the visibility regions of radius $\Delta$. The overlap of those regions is the penetration distance, labeled $\Psi$.  $\beta$ is visualized as the length of the solid green line between two vertices. The two orange lines are highlighted to demonstrate that $dist(v_1, v_3) = dist(v_1, v_2) + dist(v_2, v_3)$}
\label{fig:discretization_size}
\end{figure}

We discretize our space using a standard $d$-cubic honeycomb pattern, which in 3D is a homogeneous grid of squares. The discretization size $\beta$ is calculated through a geometric formula that leverages the known properties of SPARS combined with the space's metric function and number of dimensions $d$. We use the sparse delta factor $\Delta$ that specifies the radius of visibility, or coverage, a vertex provides over the c-space. We introduce the concept of \textit{penetration distance} $\Psi$ that defines the amount of overlap between two neighboring vertex's visibility regions. $\Psi$ should be some small value greater than zero, with the trade off that that the smaller the value the lower the probability of an edge being created between two vertices with a shared \emph{interface}, but the larger the value the more vertices are required in the graph. Here, an \textit{interface} $i(v_1, v_2)$ between two vertices $v_1$ and $v_2$ is the shared boundary of their visibility regions, as defined in \cite{visibility_prm}. This is illustrated in Figure \ref{fig:discretization_size} where the vertices $v$ are each configurations of two dimensions with values $(x,y)$.

We desire to find a value for $\beta$ that distributes the vertices such that, lacking any constraints, would provide complete coverage for a given $\Delta$ across the c-space. To achieve this, we must find the maximum distance across any two discretized vertices that share an interface. Because we are using the $L_1$-norm, the max distance in 2D is:

\begin{equation} \label{eq:max_dist_l1}
\begin{split}
dist_{max} & = distL_1(v_1,v_3) \\
           & = (x_3 - x_1) + (y_3 - y_1) \\
           & = 2 \cdot \beta \\
           & = d \cdot \beta
\end{split}
\end{equation}

The last line generalizes the result to $d$ dimensions, for example the 3D case demonstrated in the right side of Figure \ref{fig:discretization_size}. To ensure complete coverage of the space, the furthest vertices with shared interfaces must have slightly overlapping visibility regions given their $\Delta$. Therefore the two vertices with a shared interface with max distance apart need to have a distance of:
\begin{equation} \label{eq:l1_delta}
dist_{max} = 2 \cdot \Delta
\end{equation}
Where the constant $2$ is invariant for all dimensions and represents the two vertices in question. For example, in Figure \ref{fig:discretization_size}a, $v_1$ and $v_3$ share an interface  and have slightly overlapping visibility regions labeled $\Psi$. 

Combining these two functions, adding the necessary $\Psi$, and solving for the discretization level:
\begin{equation} \label{eq:discretization}
\begin{split}
\beta & =  \frac{2 \cdot \Delta}{d} - \Psi \\
\end{split}
\end{equation}
that is, for a given $\Delta$, what $\beta$ to use to have the minimum number of vertices and edges in the graph. As an aside, in our experiments we also used the $L_2$-norm for testing, which requires a slightly different formula based on the ubiquitous $L_2$ distance function:
\begin{equation} \label{eq:max_dist_l2}
\begin{split}
dist_{max} & = distL_2(v_1,v_3) \\
           & = \sqrt{(x_3 - x_1)^2 + (y_3 - y_1)^2} \\
           & = \sqrt{d \cdot \beta^2} \\
\end{split}
\end{equation}
Which, similar to the $L_1$ version, results in a discretization size:
\begin{equation} \label{eq:discretization_l2}
\beta =  \sqrt{\frac{   4 \cdot \Delta^2   }{d}} - \Psi
\end{equation}

In the rest of the paper we will continue to assume the $L_1$ norm is used. It is desirable to generalize to $L_1$ because many intuitive assumptions about graph connectivity are invalidated, allowing many redundant edges to be eliminated in a roadmap. This is because it is possible for two vertices that share an interface with each other (but no edge) to have the same length path via another vertex than by a direct edge between them.

\subsection{Exact method for choosing $t$-stretch factor}
\label{sec:tstretch}

\begin{figure}[!htb]
\centering
\includegraphics[width=3.4in]{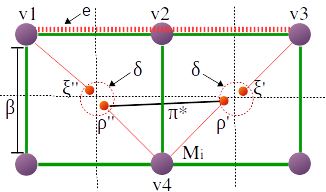}
\caption{Demonstration of geometric method for choosing $t$-stretch factor. The red dashed line is the candidate edge that should be avoided from being added. The orange vertices $\xi, \rho$ are the interface nodes that represent an interface between the vertices $[v_1, v_2]$. The black dashed lines represent the exact interfaces between the vertices.}
\label{fig:t_stretch}
\end{figure}

The SPARS algorithm path quality guarantees are largely based on the $t$-stretch factor. The value of $t$ critically affects the number of edges added to the graph and thus the performance of search. Rather than arbitrarily choose a value, we develop a formula to calculate this stretch factor to remove redundant edges added by the default SPARS method. The geometry we chose to avoid is the overlap of ``double'' edges, that is edges that span the length of three vertices instead of two as pictured in Figure \ref{fig:t_stretch}. In this Figure we show that candidate edge $e = L(v_1, v_3)$ duplicates the two adjoining edges $L(v_1, v_2)$, $L(v_2, v_3)$. If the stretch factor is too small, the SPARS quality criteria algorithm will add duplicate edges like this one. That criteria, as explained in \cite{spars2}, will add an edge if:
\begin{equation} \label{eq:tstretch_criteria}
t \cdot \pi^* < M_i
\end{equation}
where $\pi^*$ is the optimal path between the interior interface nodes $(\rho', \rho'')$ and $M_i$ is the midpoint path $[\rho'', v_4, \rho']$. We want to find the worst-case shortest length of $\pi^*$, which is when each pair of representative nodes $[\xi', \rho']$ are at their maximum distance apart. This length is $\delta$ by definition. Since we know the discretized distance $\beta$ between vertices and that we are using the $L_1$ metric function we can solve for these two line segment lengths:
\begin{equation} \label{eq:tstretch_criteria2}
\begin{split}
            M_i & = \frac{1}{2}(2\beta+2\beta) \\
                & = d\beta \\
          \pi^* & = \beta - 4\delta \\
\end{split}
\end{equation}
Solving Equation \ref{eq:tstretch_criteria} with Equation \ref{eq:tstretch_criteria2} gives us our minimum $t$-stretch factor to prevent these double edges from occurring:
\begin{equation} \label{eq:stretch}
t > \frac{d \beta}{\beta - 2\delta}
\end{equation}

\vspace{1mm}
\subsection{Methods to Reduce Outdated/Redundant Edges}

Often during roadmap construction edges are added to the graph based on the interface or quality criteria that are later no longer needed after newer vertices have changed the visibility regions of the c-space. Two techniques were added in Bolt to remove these unnecessary edges:  

First, a simple delay is added in utilizing the SPARS quality criteria until the c-space has a high probability of having full vertex coverage. This means most, if not all, of the free c-space is within the $\Delta$ visibility region of a vertex. In practice random samples are added without checking against the quality criteria until some number of failures $M_{coverage}$ occur. In our testing we used 5000 failures. After this threshold is reached, we continue to add random samples but with the quality criteria enabled. 

Second, we clear all nearby edges within a $\Delta$ radius of any new vertex added. This causes all edges in that visibility region to be re-created taking into account the new vertex, which results in some previous edges never being recreated.

\subsection{Improving Connectivity Criteria}

In the original SPARS implementation we found that more vertices were being added to satisfy the connectivity criteria than necessary. Whenever a new sample $q$ was able to connect at least two nodes $w_1, w_2$ within distance $\Delta$ that are otherwise disconnected, the sample $q$ is added to the graph with corresponding edges to the disconnected neighbors $w_1, w_2$. While this is sometimes necessary due to constraints, often an edge $e_{direct}$ can be added that directly connects $w_1, w_2$ and avoids adding $q$. This reduces the number of overall vertices in the graph. This new edge $e_{direct}$ still upholds the property that all $\forall e \in E_g : |e| \leq 2\Delta$ because all searched neighboring nodes of new sample $q$ are of distance $\leq \Delta$. Therefore two neighbors of $q$ on opposite extremums of its visibility region are at most $2\Delta$ apart.


\subsection{Improving the Smoothed Quality Path Criteria}
\label{label:quality2}
\begin{figure}[!htb]
\centering
\includegraphics[width=3.4in]{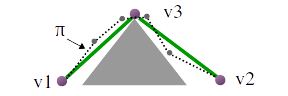}
\caption{Example of a smoothed path $\pi$ that does not improve the path length between $v_1$ and $v_2$ but only increases the graph size when added.}
\label{fig:quality_path_bug}
\end{figure}
An improvement is presented that alters the logic for connecting two vertices $[v_1, v_2]$ for the quality criteria. If the connecting edge $e \in L(v_1, v_2)$ is found necessary, but no direct edge can be added due to obstacles, the SPARS algorithm will create a new path with configurations supporting the interfaces $i(v_1,v_3)$ and $i(v_3,v_2)$ and then ``try to smooth the remaining path as much as possible'' \cite{spars_journal}. However, there are often cases, especially in $L_1$, where the current graph already has the optimal path around the obstacles and the new smooth path does not improve path length between $[v_1, v_2]$, as shown in Figure \ref{fig:quality_path_bug}. As proposed in SPARS, the graph will still add unnecessary nodes and edges from the smoothed path, increasing the size of the graph. In our improvement, we add an extra check requiring that the distance of the newly smoothed path be less than the current shortest path between the two vertices. This is accomplished with an additional call to A*.

\subsection{Modification of Quality Criteria for $L_1$ Space} \label{sec:quality}

Here we partially relax the SPARS requirement in $L_1$ spaces that all interfaces for pairs of vertices $v_1, v_2 \in V_S$ are eventually connected by an edge as the number of samples approach infinity. The result is that superfluous edges are avoided from being added. This is implemented by running A* to find the shortest path between two vertices during the quality criteria checks. 

\begin{figure}[!htb]
\centering
\includegraphics[width=3.4in]{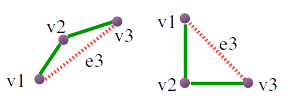}
\caption{Two examples of sets of 3 vertices with monotonic values in multiple dimensions. In both examples the addition of edge $e_3$ does not improve the path length between $v_1$ and $v_3$ with a $L_1$ metric function.}
\label{fig:monotonic}
\end{figure}

This relaxation applies to sets of three vertices whose values in each dimension are monotonic. Formally: for all sets of vertices ${v_1, v_2, v_3} \in V_G$ where edge $e_1 = L(v_1, v_2) \in E$ and $e_2 = L(v_2, v_3) \in E$, for every dimension $x$ in $C$ the set of values $[x_1, x_2, x_3]$  all increase or decrease in the same direction. When this geometry is present, the original SPARS algorithm will eventually add a third edge $e_3$ as part of its \textit{quality} criteria. However, it can be easily shown that, in an $L_1$ space, $e_3$ has the same length as the two other edges $e_1$, $e_2$ added together, as shown in Figure \ref{fig:monotonic}. Therefore, in terms of path length this third edge does not improve the quality of paths being generated in the graph. This relaxation breaks the SPARS proof \textit{asymptotic near-optimality w/additive cost} which we revise in section \ref{label:proof}.

This modification is implemented within the quality criteria tests. First, within the \texttt{Add\_Shortcut} function of SPARS \cite{spars_journal} we add an additional condition that requires the length of the candidate edge $L(v_1,v_2)$ be \textit{greater than} the length of the shortest path $\pi_G$ in $G_s$ between $v_1$ and $v_2$. With a $L^2$-norm distance function this would be impossible since no edge currently exists between $v_1$ and $v_2$ in the \texttt{Add\_Shortcut} function. However in an $L^1$-norm space this is guaranteed to happen whenever a discretized lattice structure of vertices is used to cover the space. This shortest path $\pi_G$ is found by running A* for every candidate quality edge.

\begin{figure*}[!t]
\centering
\includegraphics[width=\textwidth]{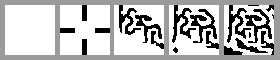}
\caption{The five 2D collision environments of sequentially more complex difficulty used for testing Bolt: Map 1 through Map 5. For the 3D case the obstacles were simply extruded into the third dimension for simplicity of debugging.}
\label{fig:collision_environments}
\vspace{1mm}
\end{figure*}

\section{Properties of Bolt} \label{label:proof}

Bolt is based on SPARS2 and has the same asymptotically near-optimal path quality guarantees. However in section \ref{sec:quality} we relaxed the requirement that all vertices that share an interface be connected by an edge in special circumstances. This breaks the \textit{Connected Interfaces} and \textit{Spanner Property} proofs in the SPARS2 paper. Next we modify those proofs from the original:
 
\begin{theorem}[Connected Interfaces - Modified]
\textit{Using the $L_1$-norm, $\forall v_1, v_2 \in V_S$ which share an interface, either $\exists L(v_1, v_2) \in E_S$ or $\exists \pi(v_1, v_2) \in G_S$ where $|\pi(v_1, v_2)| = |L(v_1, v_2)|$ with probability 1 as $M$ goes to infinity.}
\end{theorem}

The base proof is found in the original SPARS2 \textit{Connected Interfaces theorem} and says that if there exists an interface between two vertices, there is a non-empty set of sampled states that will eventually be generated that will bridge the interface with a new edge. This is guaranteed by the SPARS2 interface criteria.

Our modification of this theorem allows interface edges to \textit{not} be added when there exists a path through $G_s$ with the same length. Because there is already a path in $G_s$ with the same length as the candidate edge, not adding the edge will not affect the path length of any returned solution in $G_s$. However it still must be proved the theoretical guarantees have not been violated in the next lemma.
 
\begin{lemma}[Coverage of Optimal Paths by $G_s$ Modified]
\textit{Consider an optimal path $\pi^*$ in $C_{free}$. The probability of having a sequence of vertices in $S$, $V_{\pi}$ = ($v_1, v_2, ..., v_n$) with the following properties approaches 1 as M goes to infinity}:
\begin{itemize}
\item $\forall q \in \pi^*(q_0,q_m)$, $\exists v \in V_\pi : L(q,v) \in C_{free}$

\item $L(q_0, v_1) \in C_{free}$ and $L(q_m, v_n) \in C_{free}$

\item $\forall v_i, v_{i+1} \in V_\pi$, $L(v_i, v_{i+1}) \in E$ or $\exists \pi_G(v_i, v_{i+1}) $ where $|\pi_G(v_i, v_{i+1})| = |L(v_i, v_{i+1})|$
\end{itemize}
\label{lemma:coverage}
\end{lemma}

Here we have modified the third bullet point in Lemma \ref{lemma:coverage} to allow for paths $\pi_G$ that are not direct edges to provide coverage for an optimal path $\pi^*$ so long as this $\pi_G$ has the same length. Again, the resulting solution path must be of the same path quality because the substitute path proposed here has the same length as the direct edge it replaces.

The remaining proofs of the original SPARS hold given our modified lemmas above because any use of the removed edges can be substituted with the $\pi_G$ of same length.

\section{Results}

\begin{figure}[!t]
\centering
\includegraphics[width=3.4in]{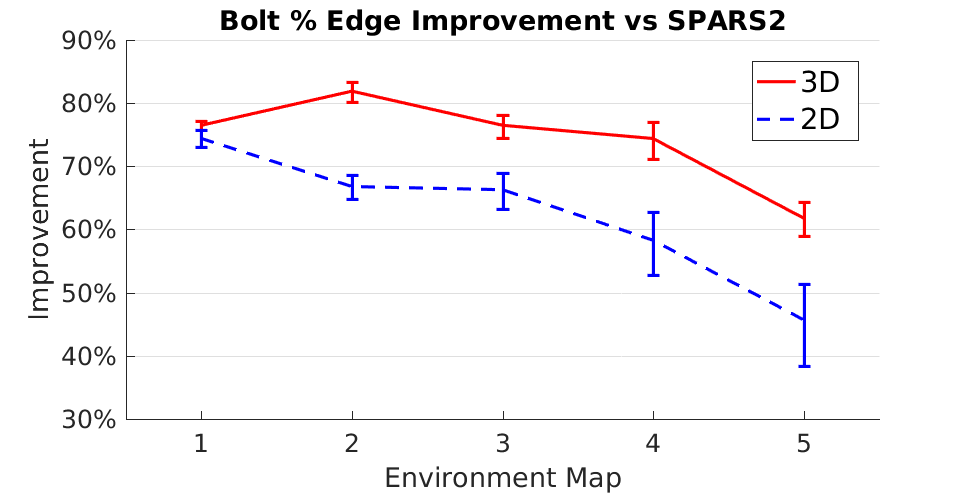}
\includegraphics[width=3.4in]{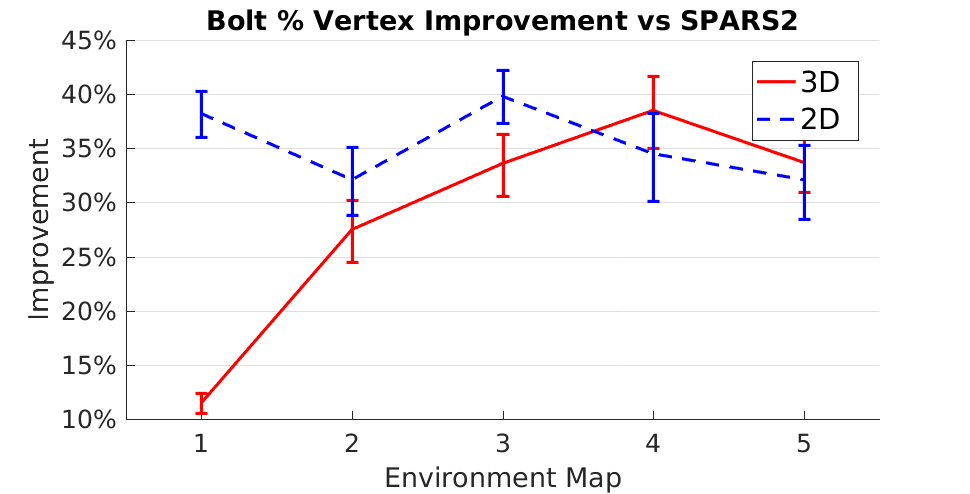}
\caption{Percent of edges (top) and vertices (bottom) improvement between SPARS2 vs Bolt roadmaps}
\label{fig:nedges}
\end{figure}

We compare the size of the graph generated by SPARS2 and Bolt across multiple environments of various complexities. The 5 environments we tested against in two and three dimensions are pictured in Figure \ref{fig:collision_environments}, ranging from a obstacle-free environment to a highly-cluttered map with many narrow passages. For each map we ran 10 trials for both SPARS2 and Bolt. Runs were tested with the stretch factor calculated as described in Section \ref{sec:tstretch} and shown in Table \ref{tab:sparseDelta}. The termination condition $M$ was set very conservative to ensure with high probability that every possible edge and vertex was added to the graph for full coverage, requiring no new edge or node be added for $M=15,000$ random samples before considering the graph "complete".
\vspace{1mm}
\begin{table}[!htbp]
\caption{Parameters of Bolt} 
\begin{center}
\begin{tabular}{ c c c c c}
d & $\Delta$ & $\delta$ & $t$ & $\Psi$\\
\hline
2 & 6.93 & 0.693 & 3.36 & 0.01\\
3 & 8.49 & 0.849 & 7.68 & 0.01\\
\end{tabular}
\label{tab:sparseDelta}
\end{center}
\end{table}

We used the SPARS2 implementation provided by the original authors in OMPL \cite{ompl}, with two minor modifications. We used an obstacle clearance as suggested in the original SPARS paper, set to 1 unit, and we fixed a bug in the quality path smoothing implementation so that the graph did not increase in size infinitely. Whenever possible we used the exact same parameters for both SPARS2 and Bolt.

For every generated roadmap we planned 1000 random paths through the space and verified the result was with the $t$-stretch factor of the optimal path. This also ensured we had sufficient coverage and connectivity between all possible states --- we had no failed plans in our tests. Each random path was then smoothed and the different in path quality recorded.

The resulting space optimizations are shown in Figure \ref{fig:nedges}. For a 2D c-space without obstacles Bolt adds 74\% fewer edges and 38\% fewer vertices. In the most cluttered environment Bolt still generates 46\% fewer edges and 32\% fewer vertices. Similar results are shown for the 3D case. Notably for collision free spaces in 3D, there is only a small improvement (11\%) with Bolt in the number of verities necessary to cover the space. This suggests our vertex count improvements work best in cluttered environments and that random sampling performs similarly in free space for both algorithms.

\begin{figure}[!t]
\centering
\includegraphics[width=3.4in]{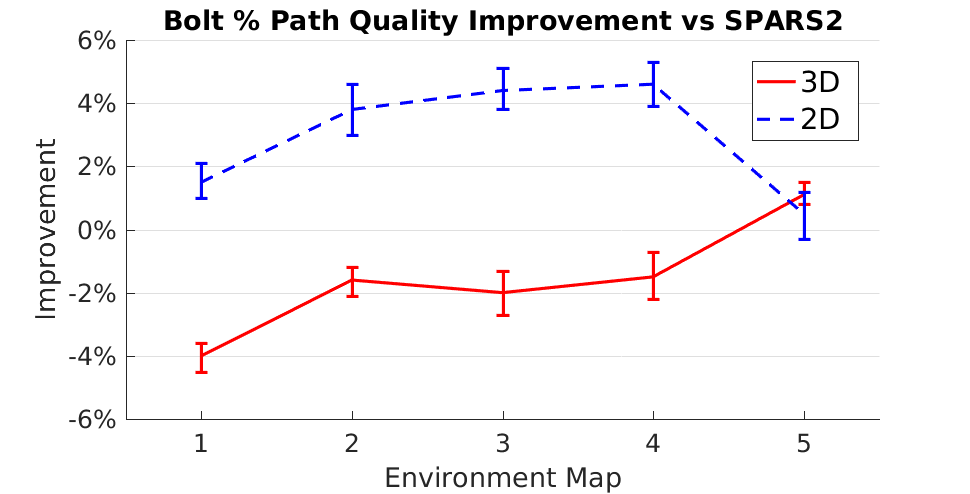}
\caption{Path quality improvement between SPARS2 vs Bolt roadmaps}
\label{fig:pathquality}
\end{figure}

The resulting path quality was compared between Bolt and SPARS in Figure \ref{fig:pathquality}. Given the standard deviations of errors, the path quality of both methods were essentially the same. However in the 2D data set Bolt returned paths with around 3\% better path quality with is impressive given the corresponding 62\% average reduction in edges. In the 3D data set Bolt had around 2\% \textit{worse} path quality, which meets expectations that the reduction in edges and verities would result in slightly worse path quality.

\begin{figure}[!t]
\centering
\includegraphics[width=3.4in]{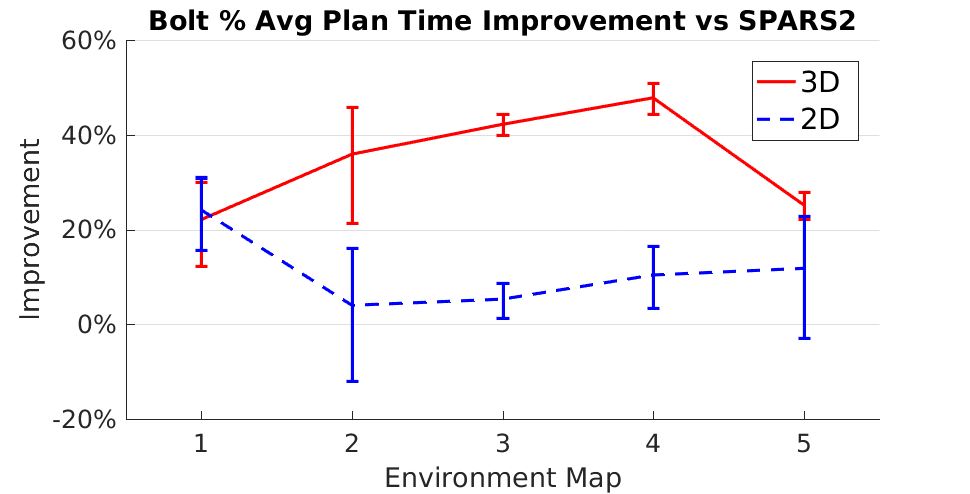}
\caption{Planning time improvement between SPARS2 vs Bolt roadmaps}
\label{fig:avgplantime}
\end{figure}

The pure A* planning time over the pre-computed graph, without changing environments or collision checking, is compared in Figure \ref{fig:avgplantime}. Bolt is faster than SPARS2 on average 11\% in 2D and 34\% in 3D, indicating higher dimensions would continue to improve online query resolution time.



Contributions of each individual improvement to the reduction of edges in SPARS2 are as follows for the 2D space using map 3. The delay in quality criteria 38\%, t-stretch factor formula 15\%, connectivity criteria check 36\%, smoothed path quality criteria 12\%, edge removal after vertex addition -2\%, edge improvement rule 27\%, discretization pre-sampling -16\%. Some of the features depend on each other for improvement, so there is actually a increase in graph size for two of the features. Notably - the removal of edges around new vertices is highly dependent on the connectivity criteria check, and upon further investigation it was found that this feature still contributes to a 2\% improvement in the number of edges in the Bolt algorithm. Similarly, the use of a discretized lattice does not improve the SPARS2 algorithm if used alone, but it overall improvement to the Bolt algorithm's average edge count was measured to be 12.45\%.

\section{Discussion}

We have shown that our improved sparse roadmaps in 3D reduce the number of edges on average 74\% and vertices 29\%. This resulted in a consistent speedup in solution time (average 35\%) and almost no loss in path quality (-2\%). 

It is difficult to compare the size savings of our method to traditional sampling-based roadmap planners such as PRM or PRM* because they lack an appropriate termination condition and therefore any results on the graph size of those planners would be arbitrary. For both Bolt and SPARS2 we are able to stop adding samples to the graph when we know with very high probability that no further samples can be added to the graph.

In future work, we wish to extend experimental validation to full 6-DOF and 12-DOF robotic arm c-spaces. While we have been successful in pre-computing sparse graphs for such systems, this takes multiple days with current hardware (given our termination criterion that all edges and verities are added), which would not allow us to systematically study the properties of the proposed algorithm. Instead we provide theoretical results for $d$-dimensions and show trends when moving from 2D to 3D. 

\section{Conclusion}

We have shown many techniques that further improve the advantages to using sparse roadmap spanners for motion planning, in particular demonstrating improvements of up to 77\% reduction in graph size. Pre-computing a full sparse roadmap for motion planning allows for more deterministic solutions to be solved faster. Expensive invariant constraints such as self collision checking are built into the roadmap ahead of time. 

Future work includes further validation in higher dimensional spaces including single and dual arm robots, and the ability to vary the discretization level for different joints. Other areas include pruning unreachable subgraphs, adding more density to areas of the roadmap that are used most often based on experience, and speeding up generation of the roadmap for very large c-spaces.


\addtolength{\textheight}{-12cm}   


\bibliographystyle{IEEEtran} 
\bibliography{bolt} 

\end{document}